\documentclass[10pt,twocolumn,letterpaper]{article}

\usepackage{wacv}
\usepackage{times}
\usepackage{epsfig}
\usepackage{graphicx}
\usepackage{amsmath}
\usepackage{amssymb}
\usepackage[accsupp]{axessibility}
\usepackage[linesnumbered,ruled,vlined,boxed]{algorithm2e}
\usepackage{multirow}
\usepackage[table,xcdraw]{xcolor}
\usepackage{pifont}
\newcommand{\cmark}{\color{blue}{\ding{51}}}
\newcommand{\xmark}{\color{blue}{\ding{55}}}

\def\wacvPaperID{1193} 

\wacvfinalcopy 

\ifwacvfinal
\fi

\ifwacvfinal
\usepackage[breaklinks=true,bookmarks=false]{hyperref}
\else
\usepackage[pagebackref=true,breaklinks=true,colorlinks,bookmarks=false]{hyperref}
\fi

\begin{document}

\newcommand{\mytitle}{\vspace{-.5in}
\textbf{\textbf{DAD}: \textbf{D}ata-free \textbf{A}dversarial \textbf{D}efense at Test Time}
}
\title{\mytitle}

\author{
\begin{tabular}{c c c}
\centering
  \hspace{0.4in}Gaurav Kumar Nayak\thanks{denotes equal contribution.} & \hspace{-0.9in}Ruchit Rawal\footnotemark[1] & \hspace{-0.9in}Anirban Chakraborty  \vspace{0.1in}\tabularnewline
  & \hspace{-0.9in}Department of Computational and Data Sciences \tabularnewline
& \hspace{-0.9in}Indian Institute of Science, Bangalore, India \tabularnewline &
\hspace{-0.9in}{\tt\small \{gauravnayak, ruchitrawal, anirban\}@iisc.ac.in}
\vspace{0.1in}
\end{tabular}\\
}
\maketitle
\ifwacvfinal
\thispagestyle{empty}
\fi
\vspace{-0.1in}
\begin{abstract}
   Deep models are highly susceptible to adversarial attacks. Such attacks are carefully crafted imperceptible noises that can fool the network and can cause severe consequences when deployed. To encounter them, the model requires training data for
adversarial training or explicit regularization-based techniques.
   However, privacy has become an important concern, restricting access to only trained models but not the training data (\eg biometric 
   data). Also, data curation is expensive and companies may have proprietary rights over it. To handle such situations, we 
   propose a completely novel problem of `\underline{test-time} adversarial defense \underline{in absence of} training data and even their statistics'. 
   We solve it in two stages: a) detection and b) correction of adversarial samples. Our adversarial sample detection framework 
   is initially trained on arbitrary data and is subsequently adapted to the unlabelled test data through unsupervised domain adaptation. We further correct the predictions on detected adversarial samples by transforming them in Fourier domain and obtaining their low frequency component at our proposed suitable radius for model prediction.
We demonstrate the efficacy of our proposed technique via extensive experiments against several adversarial attacks and for different model architectures and datasets. 
For a non-robust Resnet-$18$ model pretrained on CIFAR-10, our detection method correctly identifies $91.42\%$ adversaries. 
Also, we significantly improve the adversarial accuracy from $0\%$ to $37.37\%$ with a minimal drop of $0.02\%$ in clean accuracy on state-of-the-art `Auto Attack' without 
having to retrain the model.

\end{abstract}

\section{Introduction}
\label{sec:intro}
Deep learning models have emerged as effective solutions to several computer vision and machine learning problems. 
However, such models have become unreliable as they may yield incorrect predictions when encountered with input data added with a carefully crafted human-imperceptible noise, also termed as `adversarial noise'~\cite{szegedy2013intriguing}. These vulnerabilities of the deep models can also cause severe implications in applications involving safety and security concerns such as biometric authentication via face recognition in ATMs~\cite{middlehurst2015china}, mobile phones~\cite{faceid} etc. It can even become life crucial in self-driving cars~\cite{kurakin2016adversarial, eykholt2018robust} where autonomous vehicles can be made to take incorrect decisions when the important traffic objects are manipulated, \eg stop signs. 

\begin{figure*}[htp]
\centering
\centerline{\includegraphics[width=\textwidth]{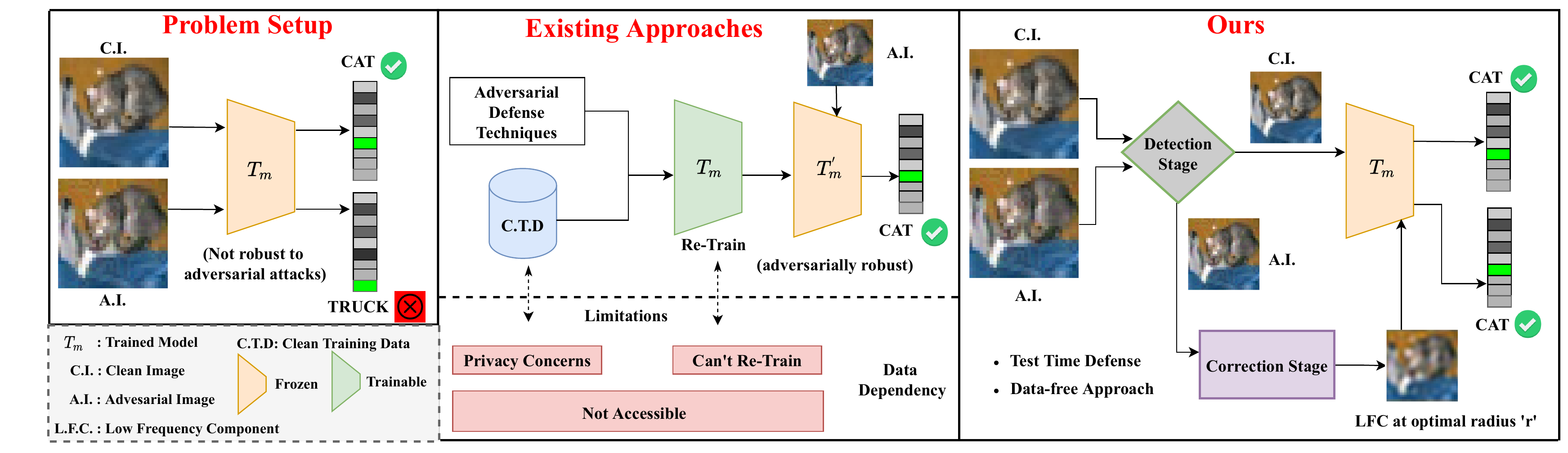}}
\caption{Comparison of existing approaches and ours. Traditional approaches fail to handle data privacy scenarios, Our method provides data-free robustness against adversarial attacks at test time without even retraining the pretrained non robust model $T_{m}$.}
\label{fig:overview}
\end{figure*}
Several attempts have been made to make the deep models robust against the adversarial attacks. We can broadly categorize them as: i.) adversarial training~\cite{goodfellow2014explaining, madry2017towards} and ii.) non-adversarial training methods~\cite{jakubovitz2018improving}. These two family of approaches have their own limitations as the former one is more computationally expensive while the latter provides weaker defense, albeit at a lower computational overhead. Instead of making the model robust, there are also approaches to detect these attacks~\cite{liang2018detecting,higashi2020detection,yang2020ml,aigrain2019detecting,xu2017feature,lorenz2021detecting,wang2019detecting,kwon2021classification}. These methods often require retraining of the network~\cite{higashi2020detection, aigrain2019detecting, kwon2021classification}. Few of them use statistical~\cite{yang2020ml, xu2017feature, wang2019detecting} and signal processing techniques~\cite{lorenz2021detecting, liang2018detecting}. 
All these existing works have a strong dependency on either the training data or their statistics. However, recently several works (\eg~\cite{nayak2019zero,yin2020dreaming,zhang2021data,yoo2019knowledge}) have identified many use cases where the training data may not be freely available, but instead the trained models. For example, pretrained models on Google’s JFT-300M~\cite{kolesnikov2020big} and Deepface model~\cite{taigman2014deepface} of Facebook do not release their training data. The training data may not be shared for many reasons such as data privacy, proprietary rights, transmission limitations, etc. 
Even biometric data are sensitive, prohibiting their distribution due to privacy. Also, several companies would not prefer sharing their precious data freely due to competitive advantage and the expensive cost incurred in data curation and annotation. Thus, we raise an important concern: \textit{‘how to make the pretrained models robust against adversarial attacks in absence of original training data or their statistics’}.

One potential solution is to generate pseudo-data from the pretrained model and then use them as a substitute for the unavailable training data. Few works attempt to generate such data either directly via several iterations of backpropagations~\cite{nayak2019zero, yin2020dreaming} or through GAN based approaches involving complicated optimization~\cite{addepalli2020degan}. 
However, the pseudo-data generation process is computationally expensive. Furthermore retraining the model on the generated data using adversarial defense techniques is an added computation overhead. This motivates an alternative strategy that involves test time adversarial detection and subsequent correction on input space (data) instead of model, without generating any computationally expensive synthetic samples. However, even identifying the adversarial input samples with no prior knowledge about the training data is a non-trivial and difficult task. After detection, we aim to go a step ahead to even correct the adversarial samples which makes the problem extremely challenging and our proposed method is an initial attempt towards it . The problem set up and the major difference between existing methods and our proposed solution strategy is also summarized in Figure~\ref{fig:overview}.

Our proposed detection method leverages adversarial detector classifier trained on any arbitrary data. In order to reduce the domain shift between that arbitrary data on which the detector is trained and our unabelled test samples, we frame our detection problem as an unsupervised domain adaptation (UDA). To further reduce the dependency on the arbitrary data, we use source-free UDA technique~\cite{liang2020we} which also allows our framework to even use any off-the-shelf pretrained detector classifier. Our correction framework is based on human cognition that human beings ignore the inputs that are outside a certain frequency range~\cite{wang2020towards}. However, the model predictions are highly associated with high frequency components \cite{wang2020high}. As adversarial attacks disrupt the model predictions, hence, we ignore the high frequency of the input beyond a certain radius. Selecting a suitable radius is crucial as removing high frequency components at low radius leads to lower discriminability while at high radius favours adversarial attacks. Adversarial noise creeps in when we aim for high discriminability. Therefore, we also propose a novel algorithm that finds a good trade off between these two. Our correction method is independent of the detection scheme. Hence, any existing data-dependent detection techniques can also benefit from our correction framework by easily plug in at test time to correct the detected adversaries. Also, as we 
do not modify the trained model ($T_{m}$ in Figure~\ref{fig:overview})
, our method is architecture agnostic and works across a range of network architectures. 

Our overall contributions can be summarized as follows:
\begin{itemize}
\itemsep0em
\item We are the first to attempt a novel problem of data-free adversarial defense at test time.
\item We propose a novel adversarial detection framework based on source-free unsupervised domain adaptation technique (Sec.~\ref{subsec:detection}), which is also the first work that 
does not depend on the training data 
and their statistics. 
\item Inspired from human cognition, our correction framework analyzes the input data in Fourier domain and discards the adversarially corrupted high-frequency regions. A respectable adversarial accuracy is achieved by selecting the low-frequency components for each unlabelled sample at an optimal radius $r^{*}$ proposed by our novel algorithm (Sec.~\ref{subsec:correction}).
\item We perform extensive experiments on multiple architectures and datasets to demonstrate the effectiveness of our method. Without even retraining the trained non-robust Resnet-18 model on CIFAR-10 and in absence of training data, we obtain a significant improvement in the adversarial accuracy from $0\%$ to $37.37\%$ with a minimal drop of $0.02\%$ in the clean accuracy on state-of-the-art Auto Attack~\cite{croce2020reliable}.
\end{itemize}
\section{Related Works}
\label{sec:related}
\subsection{Adversarial Detection}
Adversarial detection methods aim to successfully detect adversarially perturbed images. These can be broadly achieved by using trainable-detector or statistical-analysis based methods. The former usually involves training a detector-network either directly on the clean and adversarial images in spatial~\cite{lee2018unified, ma2018lid, jan2017detect} / frequency domain~\cite{Harder2021SpectralDefenseDA} or on logits computed by a pre-trained classifier~\cite{jonathan2019introspect}. Statistical-analysis based methods employ statistical tests like maximum mean discrepancy~\cite{kathrin2017detect} or propose measures~\cite{reuben2017detect, bin2021detect} to identify perturbed images.

It is impractical to use the aforementioned techniques in our problem setup as they require training data to either train the detector (trainable-detector based) or tune hyper-parameters (for statistical-analysis based). We tackle this by formulating the detection of adversarial samples as a source-free unsupervised domain adaptation setup wherein we can adapt a detector trained on arbitrary (source) data to unlabelled (target) data at test-time.
\subsection{Adversarial Robustness}
While numerous defenses have been proposed to make a model robust to adversarial perturbations, adversarial training is arguably the only one that has stood the test of time. Szegedy \etal~\cite{szegedy2013intriguing} first articulated adversarial training as augmenting the training data with adversaries to improve the robustness performance to a specific attack. Madry \etal~\cite{madry2017towards} proposed the Projected Gradient Descent (PGD) attack, which when used in adversarial training provided robustness against a wide-class of iterative and non-iterative attacks. Apart from adversarial-training, other non-adversarial training based approaches primarily aim to regularize the network to reduce overfitting and achieve properties observed in adversarially trained models explicitly. Most notably, Jacobian Adversarially Regularized networks~\cite{chan2019jacobian} achieve adversarial robustness by optimizing the model’s jacobian to match natural training images.

Since we aim to make the model robust at test-time, it’s not possible to apply either adversarial training or regularization approaches. We instead focus on reducing adversarial contamination in input data itself to achieve decent adversarial accuracy without dropping clean accuracy.
\subsection{Frequency Domain}
Wang \etal~\cite{wang2020high} in their work demonstrated that unlike humans, CNN relies heavily on high-frequency components (HFC) of an image. Consequently, perturbations in HFC cause changes in model prediction but are imperceptible to humans. More recently, Wang \etal~\cite{wang2020towards} showed that many of the existing adversarial attacks usually perturb the high-frequency regions and proposed a method to measure the contribution of each frequency component towards model prediction. Taking inspiration from these recent observations we propose a novel detection and correction module that leverages frequency domain representations to improve adversarial accuracy (without plunging clean-accuracy) at test-time. The next section explains important preliminaries followed by proposed approach in detail.
\section{Preliminaries}
\label{sec:prelims}
\textbf{Notations}: The target model $T_{m}$ is pretrained on a training dataset $D_{train}$. The complete target dataset $D_{target} = \{D_{train}, D_{test}\}$. We assume no access to $D_{train}$ but trained model $T_{m}$ is available. 
$D_{test}=\{x_{i}\}_{i=1}^{N}$ is unlabelled testing data containing $N$ test samples. We denote a set of adversarial attacks by 
$A_{attack}= \{A_{j}\}_{j=1}^{K}$ where K is the number of different attacks. The $i^{th}$ test sample $x_{i}$ is perturbed by any attack $A_{j} \in A_{attack}$ that fools the network $T_{m}$ and the corresponding adversarial sample is $x_{i}^{'}$. 

We denote any arbitrary dataset by $D_{arbitrary}=\{(x_{iA}, y_{iA})\}_{i=1}^{M}$ that has $M$ labelled samples and the dataset $D_{arbitrary}$ is different from $D_{target}$. The model $S_{m}$ is trained on arbitrary dataset $D_{arbitrary}$. The adversarial sample corresponding to $x_{iA}$ is $x_{iA}^{'}$ which is obtained when the trained model $S_{m}$ is attacked by any attack $A_{j} \in A_{attack}$. 

$T_{m}(x_{i})$ and $S_m(x_{iA})$ are the logits predicted for $i^{th}$ sample from $D_{test}$ and $D_{arbitrary}$ respectively. The softmax function is denoted by $soft()$. The label predicted by network $T_{m}$ and $S_{m}$ on $i^{th}$ sample is denoted by $label(T_{m}(x_{i}))=\mathrm{argmax}(soft(T_{m}(x_{i})))$ and $label(S_{m}(x_{iA}))=\mathrm{argmax}(soft(S_{m}(x_{iA})))$. Let $A_{test}$ and $A_{arbitrary}$ are the complete set of adversarial samples that fools the model $T_{m}$ and $S_{m}$ respectively, such that $x_{i}^{'} \in A_{test}$ and $x_{iA}^{'} \in A_{arbitrary}$ for an $i^{th}$ image. The set of layers that are used for adversarial detection are denoted by $L_{advdet}$.

The fourier transform and inverse fourier transform operations are denoted by $F(.)$ and $F^{-1}(.)$ respectively. The frequency component of an $i^{th}$ sample is denoted by $f_{i}$. The low frequency component (LFC) and high frequency component (HFC) of a sample $f_{i}$ which are separated by a radius ($r$) are denoted by $fl_{ir}$ and $fh_{ir}$ respectively.

\textbf{Adversarial noise}: Any adversarial attack $A_{j} \in A_{attack}$ fools the pretrained network $T_{m}$ by changing the label prediction. An attack $A_{j}$ on an $i^{th}$ image $x_{i}$ computes an adversarial image $x_{i}^{'}$ such that $label(T_{m}(x_{i})) \neq label(T_{m}(x_{i}^{'}))$. To obtain $x_{i}^{'}$, the $i^{th}$ image $x_{i}$ is perturbed by an adversarial noise $\delta$ which is imperceptible such that $\left\lVert \delta \right\rVert$ is within some $\epsilon$. We restrict to perturbations within the $l_{\infty}$ ball of radius $\epsilon$.

\begin{figure*}[htp]
\centering
\centerline{\includegraphics[width=0.99\textwidth]{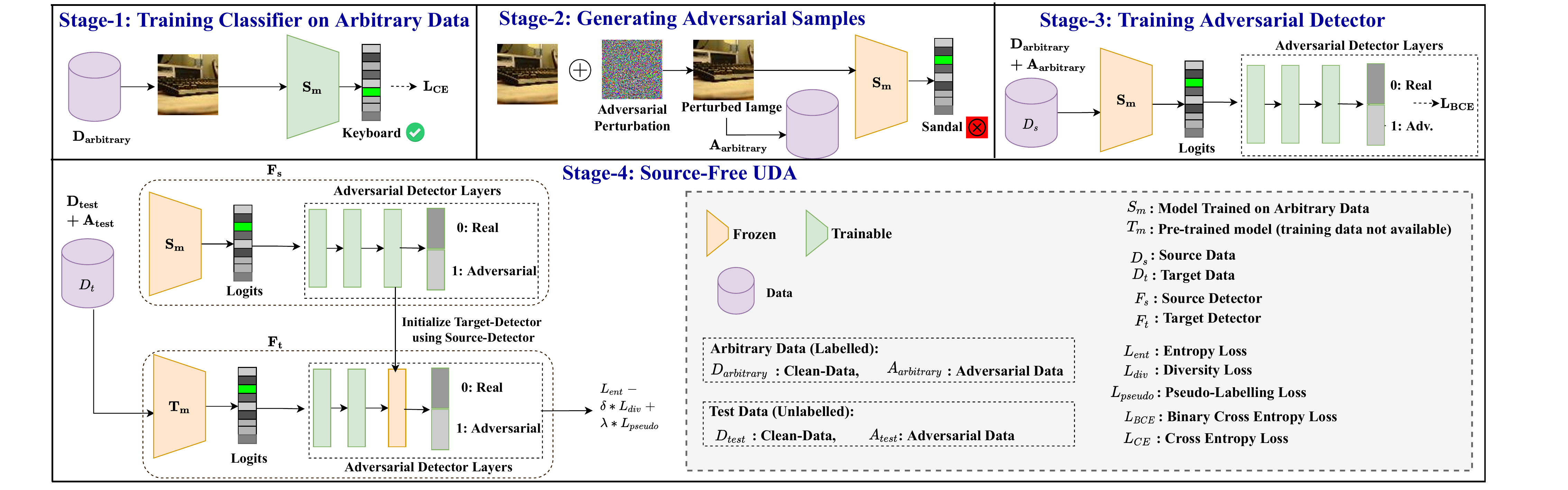}}
\caption{Our proposed detection module depicting different operations in each of the stages. We formulate adversarial detection as a source-free UDA problem where we adapt the source detector trained on arbitrary data to target detector using unlabelled clean and adversarial test samples.
}
\label{fig:detection}
\end{figure*}
\textbf{Unsupervised Domain Adaptation (UDA)}: A classifier $F_{s}$ is trained on labelled source dataset $D_{s}$ which comes from distribution $S$. The unlabelled target dataset $D_{t}$ belongs to a different distribution $T$. UDA methods attempt to reduce the domain gap between $S$ and $T$ with an objective to obtain an adapted classifier $F_{t}$ using $F_{s}$ which can predict labels on the unlabelled samples from $D_{t}$. 
If we assume $D_{s}$ to be unavailable for adaptation then this problem is referred to as source-free UDA~\cite{liang2020we, li2020model, kurmi2021domain}.

\pagebreak
\textbf{Fourier Transform (FT)}: This operation is used in image processing to transform the input image from spatial to frequency domain~\cite{bracewell1986fourier}. For any $i^{th}$ image $x_{i}$,  its FT is defined as $F(x_{i}) = f_{i}$. 
At radius $r$, 
\begin{equation}
\begin{aligned}
    fl_{ir} = LFC(f_{i}, r)\\
    fh_{ir} = HFC(f_{i}, r)
\end{aligned}
\end{equation}
The inverse of FT helps to get back the spatial domain from the frequency domain. Thus, we have:
\begin{equation}
\begin{aligned}
 xl_{ir} = F^{-1}(fl_{ir}) \\
 xh_{ir} = F^{-1}(fh_{ir})
\end{aligned}
\end{equation}
where $xl_{ir}$ and $xh_{ir}$ are the LFC and HFC of $i^{th}$ image in the spatial domain.

\section{Proposed Approach}
\label{sec:proposed}
\textbf{Test-time Adversarial Defense Set up}: Given a pretrained target model $T_{m}$, our goal is to make the model robust against the set of adversarial attacks $A_{attack}$ i.e. $T_{m}$ should not change it’s prediction on the set of adversarial samples $A_{test}$. The access to the dataset $D_{train}$ is restricted due to data privacy. The objective is to maximize the performance of $T_{m}$ on $A_{test}$ without compromising much on $D_{test}$. As shown in Figure~\ref{fig:overview}, we add a detection block before feeding the input to model $T_{m}$. The test samples which are detected as adversarial are passed to the correction module to minimize the adversarial contamination. The clean detected samples as well as corrected adversarial samples (output after correction stage) are then fed to pretrained model $T_{m}$ to get the predictions. Next we explain in detail about our proposed detection and correction modules.

\subsection{Detection module}
\label{subsec:detection}
The key idea is that if we have access to an adversarial detector which can classify samples from an arbitrary dataset as either clean and adversarial, then this binary classifier can be treated as a source classifier. The adversarial detector for the target model $T_{m}$ can be thought of as target classifier with the target data being the collection of unlabelled test samples and their corresponding adversarial samples. Thus, as described in preliminaries of UDA, we can formulate the adversarial detection as a UDA problem where:\\
$F_{s} \leftarrow$ model $S_{m}$ appended with detection layers $L_{advdet}$\\
$D_{s} \leftarrow$ mix of $D_{arbitrary}$ (clean) and $A_{arbitrary}$ (adversarial)\\
$D_{t} \leftarrow$ mix of $D_{test}$ (clean) and $A_{test}$ (adversarial)\\
$F_{t} \leftarrow$ model $T_{m}$ appended with detection layers $L_{advdet}$\\
\newline
The goal of our detection module is to obtain the model $F_{t}$ via UDA technique (to reduce the domain shift between $D_{s}$ and $D_{t}$) that can classify samples from $D_{t}$ as either clean or adversarial. \par Our detection module consists of four stages which are shown in detail in Figure~\ref{fig:detection}. Each stage performs a specific task which are described below:\\
\textbf{Stage-1}: Train model $S_{m}$ with labelled data $D_{arbitrary}$ by minimizing the cross entropy loss $\min \sum_{i=1}^{M} L_{ce}(S_{m}(x_{iA}), y_{iA})$.\\
\textbf{Stage-2}: Generate a set of adversarial samples ($A_{arbitrary}$) by using any adversarial attack $A_{j}$ from a set of adversarial attacks ($A_{attack}$) such that it fools the trained network $S_{m}$.\\
\textbf{Stage-3}: Train the adversarial detection layers ($L_{advdet}$) with input being the logits of network $S_{m}$ and output is the soft score for the two classes (adversarial and clean). Similar to \cite{jonathan2019introspect}, the layers in $L_{advnet}$ are composed of three fully connected layers containing 128 neurons and ReLU activations (except the last-layer). The first-two layers are followed by a dropout-layer with 25\% dropout-rate. Additionally, the second (after-dropout) and third layer are followed by a Batch Normalization and Weight Normalization layer respectively. The training is done using binary cross entropy loss $L_{BCE}$ on data $D_{s}$. The label smoothing~\cite{muller2019does} is also done to further improve the discriminability of the source model $F_{s}$. \\
\textbf{Stage-4}:  Perform UDA  with the source network as $F_{s}$ and the target network as $F_{t}$ where the source data is $D_{s}$ and target data is $D_{t}$. If we remove the dependency on the dataset $D_{s}$, then it can also easily facilitate any off-the-shelf pretrained detector classifier to be used in our framework and Stages $1$ to $3$ can be skipped in that case. Thus, to make our framework completely data-free and reduce its dependency on even arbitrary data, we adopt source-free UDA. To the best of our knowledge, the UDA setup for adversarial detection has not been explored previously even for data-dependent approaches. 

The target network $F_{t}$ is trained on the unlabelled dataset $D_{t}$ where network $T_{m}$ is frozen and entire layers of $L_{advdet}$ of $F_{t}$ is kept trainable expect the last classification layer and are initialized with weights of $L_{advdet}$ of $F_{s}$.  Inspired by~\cite{liang2020we}, we also use three losses for training. We minimize the entropy loss ($L_{ent}$) to enforce that the network $F_{t}$ predicts each individual sample with high confidence i.e. strong predictions for one of the classes (adversarial or clean). But this may result in a degenerate solution where only the same class gets predicted (i.e. same one-hot embedding). So, to avoid that we maximize diversity loss ($L_{div}$) that ensures high entropy across all the samples i.e. enforcing the mean of network predictions to be close to uniform distribution. Still, the unlabelled samples in $D_{t}$ can get assigned a wrong label even if the ($L_{ent}-L_{div}$) is minimized. To overcome this, pseudo-labels are estimated via self-supervision~\cite{caron2018deep,liang2020we}. The initial centroid for a class is estimated by weighted mean of all samples where the individual sample weight is the predicted score of that class. Then the cosine similarity of the samples with respect to each centroid is calculated and the label of the nearest centroid is used as initial pseudo-labels. Next, the initial centroids and pseudo-labels are updated in a similar way as in K-means. $L_{pseudo}$ is cross entropy loss on $D_{t}$ where the assigned pseudo-labels are used as ground truth labels. Thus, overall loss $L = (L_{ent}-\delta L_{div}+ \lambda L_{pseudo}$) is minimized where $\delta$ and $\lambda$ are hyperparameters. 

\subsection{Correction module}
\label{subsec:correction}
The input images which are classified as adversarial by the trained adversarial detector model $F_{t}$ are passed to this module. Here the objective is to minimize the contamination caused by adversarial attack on a given image such that the pretrained model $T_{m}$ would retain its original prediction. In other words, an $i^{th}$ sample $x_{i}^{'} \in A_{test}$ should be corrected as $x_{ic}^{'}$ so that $label(T_{m}(x_{ic}^{'})) = label(T_{m}(x_{i}))$.

The trained model gets fooled when its prediction gets altered on adversarial samples. Also, the deep model’s prediction is highly associated with the HFC of an input image~\cite{wang2020high}. To lessen the effect of adversarial attack, we need to get rid of the contaminated HFC. This also resonates with human cognition as humans use LFC for decision making while ignoring the HFCs~\cite{wang2020towards}. Consequently, the high frequency components are required to be ignored at some radius $r$. Thus, for an $i^{th}$ adversarial image $x_{i}^{'}$ (refer FT operations in  Preliminaries Sec.~\ref{sec:prelims}), we have: 
\begin{equation}\label{eq3}
\begin{gathered}
f_i' = F(x_{i}^{'})\\
\text{At any radius $r$}: fl_{ir}^{'} = LFC(f_{i}^{'}, r) ,  fh_{ir}^{'} = HFC(f_{i}^{'}, r)  \\
xl_{ir}^{'} = F^{-1}(fl_{ir}^{'})
\end{gathered}
\end{equation}
However, $xl_{ir}^{'}$ obtained on a random selection of a radius $r$  may yield poor and undesired results (refer Table~\ref{tab:ablation}). LFC’s at high radius favours more discriminability (Disc) but high adversarial contamination (AdvCont). On the other hand, LFC’s at small radius allows low AdvCont but low Disc. Hence, careful selection of suitable radius $r^{*}$ is necessary to have a good trade off between Disc and AdvCont i.e. low AdvCont but at the same time having high Disc.  Hence, the corrected $i^{th}$ image in the spatial domain at radius $r^{*}$ is given as:
\begin{equation}\label{eq4}
\begin{aligned}
x_{ic}^{'} =  xl_{ir^{*}}^{'} = F^{-1}(fl_{ir^{*}}^{'})  
\end{aligned}
\end{equation}
Estimating the radius $r^{*}$ is not a trivial task especially when no prior knowledge about the training data is available and the value of suitable $r^{*}$ can even vary for each adversarial sample. To overcome this problem, our proposed correction method estimates $r^{*}$ for each sample and returns the corrected sample which is explained in detail in Algorithm~\ref{algo_corrections}. We define the minimum and maximum radius i.e. $r_{min}$ and $r_{max}$ (Line~\ref{initial}). As discussed, there are two major factors associated with the radius selection: Disc and AdvCont. Thus, we measure these two quantities present in LFC at each radius between $r_{min}$ and $r_{max}$ with a step size of 2. We quantify the Disc using SSIM~\cite{wang2004image} which is a perceptual metric. Specifically, the Disc score for an $i^{th}$ adversarial sample (Line~\ref{disc_score}) is given as:
\begin{equation}\label{eq5}
\begin{aligned}
Disc_{score}(x_{i}^{'}, xl_{ir}^{'}) = SSIM(x_{i}^{'}, xl_{ir}^{'})
\end{aligned}
\end{equation}
\vspace{-0.1in}
\begin{algorithm}[htp]
{\small

\caption{Correction of adversarial samples ($A_{test}$) by determining the best radius $r^{*}$}
\label{algo_corrections}
\SetAlgoLined
\SetKwInOut{Input}{Input}  
\Input{Pretrained model $T_{m}$, \\
$x_{i}^{'}$: $i^{th}$ adversarial sample
}
\SetKwInOut{Output}{Output}  
\Output{$x_{ic}^{'}$ : corrected $i^{th}$ adversarial sample}
Obtain prediction on the $i^{th}$ adversarial sample: $\newline$ $advpred \leftarrow label(T_{m}(x_{i}^{'}))$\\
Initialize: $\newline$
$r_{min} \leftarrow 2$, $r_{max} \leftarrow 16$, $count \leftarrow 10$, $r^{*} \leftarrow r_{min}$\\ \label{initial}
Set dropout in training mode for $T_{m}$\\ \label{drop}
\For{$r=r_{min};\ r \le r_{max};\ r = r + 2$}{
Obtain LFC $xl_{ir}^{'}$ for the $i^{th}$ adversarial sample $x_{i}^{'}$ using equation~\ref{eq3} \\
Compute discriminability score for $xl_{ir}^{'}$: $\newline$ $Disc\_xl_{ir}^{'} \leftarrow Disc_{score}(x_{i}^{'}, xl_{ir}^{'})$ using equation~\ref{eq5}\\ \label{disc_score}
Initialize the label change rate: $lcr_{r}=0$\\ \label{initial_lcr}
\For{$k=1:count$}{
$advpred_{r} = label(T_{m}(xl_{ir}^{'}))$\\
\If{$advpred_{r} \ne advpred$}
{
$lcr_{r} = lcr_{r} + 1$
}
}\label{lcr_end}
$AdvCont\_xl_{ir}^{'} \leftarrow AdvCont_{score}(xl_{ir}^{'})= (count-lcr_{r}) / count$\\ \label{normalize}
\eIf{($Disc\_xl_{ir}^{'} - AdvCont\_xl_{ir}^{'}) > 0$}{ \label{max_r_start}
 $r^{*} = r$
}{
break;
}
}\label{max_r_end}
Obtain LFC at best radius $r^{*}$: $\newline$ $x_{ic}^{'} = xl_{ir^{*}}^{'} = F^{-1}(LFC(F(x_{i}^{'}), r^{*}))$ \label{best_r}
}
\end{algorithm}

As adversarial samples are perceptually similar to clean samples, hence SSIM score should be high to have high discriminability. We normalize SSIM between $0$ to $1$. Although increasing the radius leads to better $Disc_{score}$, it also allows the adversarial perturbations (usually located in HFC regions) to pass through. Thus, we also need to quantify AdvCont i.e. how much perturbations have crept in the LFC with respect to adversarial image $x_{i}^{'}$. To solve this, we compute the label-change-rate (LCR) at each radius. The key intuition of our method lies in the fact that if enough perturbations have passed through at some radius $r$, the adversarial component would be the dominating factor resulting in the low-pass spatial sample $xl_{ir}^{'}$ to have the same prediction as $x_{i}^{'}$. To get a better empirical estimate we compute the low-pass predictions repeatedly, after enabling the dropout. 
Enabling dropout perturbs the decision boundary slightly. Thus if adversarial noise has started dominating, the shift in decision boundary will have no/very-less effect on the model’s prediction. To quantify this effect, the LCR (at a particular radius $r$) captures the number of times the LFC prediction differs w.r.t original adversarial sample (Lines~\ref{initial_lcr}-~\ref{lcr_end}) i.e. higher LCR implies low adversarial contamination and vice-versa. 
Line~\ref{normalize} enforces max LCR (i.e. min adversarial score, denoted by $AdvCont_{score}$) as $0$ and min LCR (i.e. max $AdvCont_{score}$) as $1$. This allows us to directly compare the adversarial and discriminability score. The optimal radius for $x_{i}^{'}$ is the maximum radius at which $Disc_{score}$ is greater than $AdvCont_{score}$ (Lines~\ref{max_r_start}-~\ref{max_r_end}). The corrected $i^{th}$ sample i.e. $x_{ic}^{'}$ (Line~\ref{best_r}) is then passed to the model $T_{m}$ to get the predictions.

\section{Experiments}
\label{sec:expt}
Different architectures such as Resnet-$18$~\cite{he2016deep} and Resnet-$34$~\cite{he2016deep} are used as the target model $T_{m}$ . Each of these architectures is trained on two different datasets i.e. CIFAR-$10$~\cite{krizhevsky2009learning} and Fashion MNIST (FMNIST)~\cite{xiao2017fashion} using the standard cross-entropy loss. CIFAR-10 is a colored dataset containing RGB images while FMNIST is grayscale. 
Once the model $T_{m}$ is trained, its corresponding training set is not used any further due to the assumption of data privacy. We perform three different types of adversarial attacks on trained model Tm, namely PGD~\cite{madry2017towards}, IFGSM~\cite{kurakin2016adversarial}, and Auto Attack~\cite{croce2020reliable} (state-of-the-art) at test time. The images are normalized between $0$ to $1$ before perturbing them to create adversarial samples.  The settings for the attack parameter are followed from \cite{vivek2020plug}. For the attack on a model trained on CIFAR, the attack parameter $\epsilon$ is taken as $8/255$. Similarly for FMNIST, the $\epsilon$ parameter is 0.2. In the case of a PGD attack, the $\epsilon_{step}$ is $2/255$ and $0.02$ while the number of iterations ($N$) is $20$ and $100$ for CIFAR and FMNIST respectively. The value of $N$ remains the same for IFGSM attack while $\epsilon_{step}$ is $\epsilon/N$ for both the datasets. We perform separate analysis for each of our proposed modules in Sec.~\ref{subsec:detection_expt} (analysis of detection module)  and Sec.~\ref{subsec:correction_expt} (analysis of correction module). 
Results for more datasets are included in supplementary.
\begin{table*}[htp]
\centering
\scalebox{0.99}{
\begin{tabular}{|c|c|c|c|c|c|c|c|c|c|c|}
\hline
\multirow{2}{*}{\textbf{Dataset}} &
  \multirow{2}{*}{\textbf{\begin{tabular}[c]{@{}c@{}}Model\\ ($T_{m}$)\end{tabular}}} &
  \multicolumn{3}{c|}{\textbf{PGD}} &
  \multicolumn{3}{c|}{\textbf{IFGSM}} &
  \multicolumn{3}{c|}{\textbf{Auto-Attack}} \\ \cline{3-11} 
 &
   &
  \textbf{Overall} &
  \textbf{Clean} &
  \textbf{Adv.} &
  \textbf{Overall} &
  \textbf{Clean} &
  \textbf{Adv.} &
  \textbf{Overall} &
  \textbf{Clean} &
  \textbf{Adv.} \\ \hline \hline 
\multirow{2}{*}{CIFAR-10} & ResNet-18 & 94.01 & 95.37  & 92.64  & 85.28  & 90.85  & 79.7  & 95.69  & 99.96  & 91.42  \\ 
                          & Resnet-34 & 92.33  & 93.35  & 91.31  & 82.13  & 94.21  & 70.05  & 94.86  & 98.24 & 91.49  \\ \hline
\multirow{2}{*}{FMNIST}  & Resnet-18 & 86.23  & 97.24  & 75.22  & 89.17  & 96.28  & 82.05  & 85.52  & 98.28  & 72.75  \\ 
& Resnet-34 & 89.65  & 99.43  & 79.87  & 90.06  & 99.51  & 80.62  & 86.37  & 98.32  & 74.41  \\ \hline
\end{tabular}
}
\caption{Proposed Detection Module Performance: Detection Accuracy (in \%) on both clean and adversarial (`Overall'), clean samples (`Clean') and adversarial samples (`Adv.') for non robust target models trained on CIFAR-10 and Fashion MNIST with different architectures (Resnet-18 and Resnet-34).}
\label{detect_table}
\end{table*}
\begin{table*}[htp]
\centering
\scalebox{0.98}{
\begin{tabular}{cclllllll}
\hline
\multicolumn{1}{|l|}{{\color[HTML]{000000} }} & \multicolumn{1}{l|}{{\color[HTML]{000000} }} & \multicolumn{1}{l|}{{\color[HTML]{000000} }} & \multicolumn{2}{c|}{{\color[HTML]{000000} \textbf{PGD}}} & \multicolumn{2}{c|}{{\color[HTML]{000000} \textbf{IFGSM}}} & \multicolumn{2}{c|}{{\color[HTML]{000000} \textbf{Auto-Attack}}} \\ \cline{4-9} 
\multicolumn{1}{|l|}{\multirow{-2}{*}{{\color[HTML]{000000} \textbf{Dataset}}}} & \multicolumn{1}{l|}{\multirow{-2}{*}{{\color[HTML]{000000} \textbf{Model}}}} & \multicolumn{1}{l|}{\multirow{-2}{*}{{\color[HTML]{000000} \textbf{\begin{tabular}[c]{@{}l@{}}Clean\\ (B. A.)\end{tabular}}}}} & \multicolumn{1}{c|}{{\color[HTML]{000000} \textbf{\begin{tabular}[c]{@{}c@{}}Adv. \\ (A. A.)\end{tabular}}}} & \multicolumn{1}{c|}{\color[HTML]{000000} \textbf{\begin{tabular}[c]{@{}c@{}}Ours \\ (A. C.)\end{tabular}}}& \multicolumn{1}{c|}{{\color[HTML]{000000} \textbf{\begin{tabular}[c]{@{}c@{}}Adv. \\ (A. A.)\end{tabular}}}} & \multicolumn{1}{c|}{{\color[HTML]{000000} \textbf{\begin{tabular}[c]{@{}c@{}}Ours \\ (A. C.)\end{tabular}}}} & \multicolumn{1}{c|}{{\color[HTML]{000000} \textbf{\begin{tabular}[c]{@{}c@{}}Adv. \\ (A. A.)\end{tabular}}}} & \multicolumn{1}{c|}{{\color[HTML]{000000} \textbf{\begin{tabular}[c]{@{}c@{}}Ours \\ (A. C.)\end{tabular}}}} \\ \hline \hline
\multicolumn{1}{|c|}{} & \multicolumn{1}{c|}{vgg-16} & \multicolumn{1}{l|}{94.00} &  \multicolumn{1}{l|}{4.64} & \multicolumn{1}{l|}{41.04 \color{blue}{(36.4$\uparrow$})} & \multicolumn{1}{l|}{22.03} & \multicolumn{1}{l|}{38.37 \color{blue}{(16.34$\uparrow$})} & \multicolumn{1}{l|}{0.00} & \multicolumn{1}{l|}{40.06 \color{blue}{(40.06$\uparrow$})} \\ 
\multicolumn{1}{|c|}{} & \multicolumn{1}{c|}{resnet18} & \multicolumn{1}{l|}{93.07} & \multicolumn{1}{l|}{0.45} & \multicolumn{1}{l|}{39.39 \color{blue}{(38.94$\uparrow$})} & \multicolumn{1}{l|}{5.9} & \multicolumn{1}{l|}{38.49 \color{blue}{(32.59$\uparrow$})} & \multicolumn{1}{l|}{0.00} & \multicolumn{1}{l|}{40.25 \color{blue}(40.25{$\uparrow$})} \\ 
\multicolumn{1}{|c|}{\multirow{-3}{*}{CIFAR}} & \multicolumn{1}{c|}{resnet34} & \multicolumn{1}{l|}{93.33} & \multicolumn{1}{l|}{0.18} & \multicolumn{1}{l|}{41.71 \color{blue}{(41.53$\uparrow$})} & \multicolumn{1}{l|}{4.76} & \multicolumn{1}{l|}{40.62 \color{blue}{(35.86$\uparrow$})} & \multicolumn{1}{l|}{0.00} & \multicolumn{1}{l|}{42.40 \color{blue}{(42.40$\uparrow$})} \\ \hline \hline
\multicolumn{1}{|c|}{} & \multicolumn{1}{c|}{vgg-16} & \multicolumn{1}{l|}{91.79} & \multicolumn{1}{l|}{0.62} & \multicolumn{1}{l|}{33.09 \color{blue}{(32.47$\uparrow$})} & \multicolumn{1}{l|}{6.43} & \multicolumn{1}{l|}{33.83 \color{blue}{(27.40$\uparrow$})} & \multicolumn{1}{l|}{0.00} & \multicolumn{1}{l|}{37.99 \color{blue}{(37.99$\uparrow$})} \\ 
\multicolumn{1}{|c|}{} & \multicolumn{1}{c|}{resnet18} & \multicolumn{1}{l|}{90.31} & \multicolumn{1}{l|}{2.95} & \multicolumn{1}{l|}{32.22 \color{blue}{(29.27$\uparrow$})} & \multicolumn{1}{l|}{7.64} & \multicolumn{1}{l|}{32.38 \color{blue}{(24.74$\uparrow$})} & \multicolumn{1}{l|}{0.00} & \multicolumn{1}{l|}{35.80 \color{blue}{(35.80$\uparrow$})} \\ 
\multicolumn{1}{|c|}{\multirow{-3}{*}{F-MNIST}} & \multicolumn{1}{c|}{resnet34} & \multicolumn{1}{l|}{90.82} & \multicolumn{1}{l|}{1.85} & \multicolumn{1}{l|}{33.23 \color{blue}{(31.38$\uparrow$})} & \multicolumn{1}{l|}{5.57} & \multicolumn{1}{l|}{33.73 \color{blue}{(28.16$\uparrow$})} & \multicolumn{1}{l|}{0.00} & \multicolumn{1}{l|}{35.82 \color{blue}{(35.82$\uparrow$})} \\ \hline
\end{tabular}
}
\caption{Notations: B.A. - Before Attack, A.A. - After Attack, A.C. - After Correction. Performance (in \%) without and with our proposed correction module across VGG and Resnet architectures on different datasets i.e. CIFAR 10 and Fashion MNIST. The non-robust trained models performs poorly on adversarial samples which is significantly recovered through our correction method by achieving major boost in adversarial accuracy. The symbol ({\color{blue}$\uparrow$}) denotes increment in performance by our method (A.C.) over (A.A.).}
\label{tab:correction}
\end{table*}

\subsection{Performance Analysis of Detection Module}
\label{subsec:detection_expt}
To evaluate our detection module we perform extensive experiments with TinyImageNet as our arbitrary dataset ($D_{arbitrary}$). The source model $F_{s}$ comprises of a ResNet-$18$ classifier ($S_{m}$) followed by a three-layer adversarial detector module ($L_{advdet}$). The stage-$1$ begins by training $S_{m}$ on $D_{arbitrary}$ with the standard cross-entropy loss and stochastic gradient descent optimizer with $1e-3$ and $0.9$ as the learning rate and momentum respectively. For stage-$2$ we obtain adversarial samples ($A_{arbitrary}$) by attacking $S_{m}$ with PGD attack, the parameters for which are described in detail in the supplementary. Stage-$3$ involves training of our adversarial detector where we train the layers ($L_{advdet}$) using $L_{BCE}$ loss. The input to the loss is predicted soft scores i.e. $L_{advdet}(S_{m}(x))$ where $x \in D_{arbitrary} \cup A_{arbitrary}$ and the ground-truth i.e. $1$ (adversarial) and $0$ (clean) respectively. Finally, to make our approach completely data-free we further reduce the dependency on even arbitrary datasets by formulating a source-free domain adaptation setup. Thus, the first three stages can be skipped given we have access to the pretrained detector on arbitrary data. To adapt the source-model $F_{s}$ on our unlabelled target-set, we train $F_{t}$ (frozen $T_{m}$ model followed by trainable $L_{advdet}$) with $L_{ent}$, $L_{div}$, $L_{pseudo}$ as described in Sec.~\ref{subsec:detection}. We use fixed values of $\delta$ and $\lambda$ as $0.8$ and $0.3$ for all our UDA-based detection experiments. We perform experiments over multiple target datasets (CIFAR-$10$, F-MNIST) and target-model classifiers i.e. $T_{m}$ (ResNet-$18$, ResNet-$34$) to demonstrate the effectiveness of our detection module. Refer the supplementary for results on other arbitrary datasets.

The experiment results are shown in Table~\ref{detect_table}. We can observe a high overall detection accuracy on a broad range of attacks, architectures ($T_{m}$), and datasets. We split the overall detection accuracy into adversarial and clean detection accuracies to better investigate the detector’s performance. Our detection setup is a binary-classification problem, the adversarial detection accuracy can be understood as the True-Positive-Rate of our detector i.e. proportion of samples that are adversarial and correctly classified. Similarly, the clean detection accuracy is the True-Negative-Rate i.e. proportion of samples that are clean and classified correctly. In our experiments, we observe that we achieve a very high True-Negative-Rate ($\approx$90\%-99\%) which is highly desirable in order to preserve the clean accuracy on $T_{m}$. The clean samples are directly passed to $T_{m}$, whereas the adversarially detected samples are first processed through the correction module that we describe 
in the next section.
\subsection{Performance Analysis of Correction Module}
\label{subsec:correction_expt}
Our correction module is training-free as we correct an incoming adversarially perturbed image by obtaining its LFC at optimal radius $r^{*}$. To reduce the computational complexity in calculating LFC using FT, we use FFT operations in the experiments. We calculate (as described in~\ref{subsec:correction}) the $AdvCont_{score}$ and $Disc_{score}$ to measure the discriminability and adversarial contamination for each LFC obtained at different radii. For obtaining the $Disc_{score}$ (eq.~\ref{eq5}), we calculate the normalized SSIM-score using the open-source implementation provided by~\cite{vainf}. Our best radius $r^{*}$ is selected as the maximum radius where $Disc_{score}$ is greater than $AdvCont_{score}$. The corrected image at $r^{*}$ is obtained using eq.~\ref{eq4} where the imaginary part of $F^{-1}$ output is discarded to enable the output to be fed to the trained model $T_{m}$ to get the predictions.

In Table~\ref{tab:correction}, we present the results for our correction module. The performance of the non-robust model $T_{m}$ on clean and adversarially perturbed data is denoted by B.A. (Before Attack) and A.A. (After Attack) respectively. Assuming ideal detector, we pass each adversarial sample through our correction module and report the performance of corrected data as A.C. (After Correction). Most notably, we achieve a performance gain of $\approx35-40\%$ on the state-of-the-art auto-attack across different architectures on multiple datasets. To further investigate the efficacy of our correction algorithm across different adversarial attacks, we also perform experiments on the widely popular PGD and IFGSM attacks, and obtain a similar boost in adversarial accuracy. 

Estimating $r^{*}$ accurately is especially important to our correction module’s performance as we assume no knowledge about the training data. We verify this intuition by performing ablations on a “Random Baseline” (R.B.) wherein $r^{*}$ is chosen randomly (within our specified range) for each sample. As shown in Table~\ref{tab:ablation}, R.B. although slightly higher than A.A., is significantly lower than A.C. that indicates the usefulness of selecting $r^{*}$ appropriately. 
\begin{table}[htp]
\centering
\scalebox{0.8}{
\begin{tabular}{|c|c|l|l|l|l|l|l|}
\hline
\multicolumn{1}{|l|}{\multirow{2}{*}{\textbf{Dataset}}} &
  \multicolumn{1}{l|}{\multirow{2}{*}{\textbf{Model}}} &
  \multicolumn{2}{c|}{\textbf{PGD}} &
  \multicolumn{2}{c|}{\textbf{IFGSM}} &
  \multicolumn{2}{c|}{\textbf{Auto-Attack}} \\ \cline{3-8} 
\multicolumn{1}{|l|}{} &
  \multicolumn{1}{l|}{} &
  \multicolumn{1}{c|}{\textbf{R. B.}} &
  \multicolumn{1}{c|}{\textbf{Ours}} &
  \multicolumn{1}{c|}{\textbf{R. B.}} &
  \multicolumn{1}{c|}{\textbf{Ours}} &
  \multicolumn{1}{c|}{\textbf{R. B.}} &
  \multicolumn{1}{c|}{\textbf{Ours}} \\ \hline \hline
\multirow{3}{*}{CIFAR} & vgg-16   & 28.29 & 41.04 & 36.22 & 38.37 & 28.51  & 40.06 \\ 
                       & resnet18 & 23.15 & 39.39 & 28.12 & 38.49 & 24.82 & 40.25 \\ 
                       & resnet34 & 23.67 & 41.71 & 25.39 & 40.62 & 21.64 & 42.4 \\ \hline\hline
\multirow{3}{*}{FMNIST} & vgg-16   & 10.95 & 33.09 & 13.84 & 33.83 & 9.64 & 37.99 \\ 
                       & resnet18 & 8.55 & 32.22 & 12.87 & 32.38 & 9.50 & 35.8 \\ 
                       & resnet34 & 9.82 & 33.23  & 10.69 & 33.73  & 9.47 & 35.82 \\ \hline
\end{tabular}
}
\caption{Ablation on radius selection: Our proposed technique for radius selection leads to significant adversarial accuracy against Random baseline (R. B.) where the radius is selected randomly. The reported baseline  performance is the mean over five trials.}
\label{tab:ablation}
\end{table}
\vspace{-0.1in}
\section{Combined Detection and Correction}
\label{combined}
\begin{figure}[b]
\centering
\centerline{\includegraphics[width=0.5\textwidth]{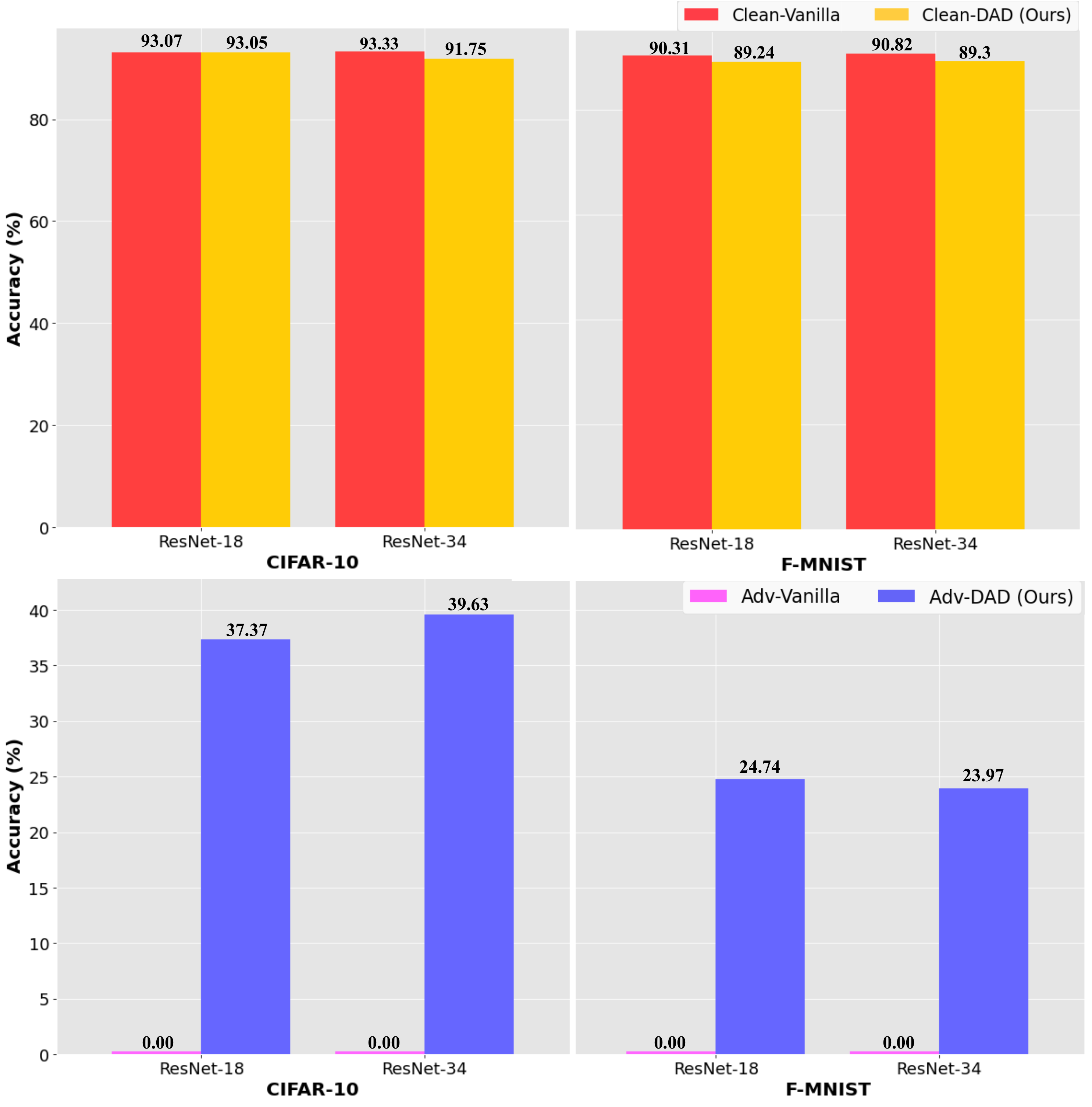}}
\caption{Performance comparison w/o (vanilla) and w/ our proposed detection and correction module to provide data-free adversarial defense (DAD) across different architectures and datasets on state-of-the-art Auto Attack.
}
\label{fig:combined}
\end{figure}
In this section, we discuss the performance on clean ($D_{test}$) and adversarially perturbed data ($A_{test}$) after combining our detection and correction modules as shown in Figure~\ref{fig:overview}. Our combined module provides \textbf{D}ata-free \textbf{A}dversarial \textbf{D}efense (dubbed as `\textbf{DAD}') at test time in an end-to-end fashion without any prior information about the training data. DAD focuses on detecting and correcting adversarially perturbed samples to obtain correct predictions without modifying the pretrained classifier ($T_{m}$) which allows us to achieve a significant gain in adversarial accuracy without compromising on the clean accuracy, as shown in Figure~\ref{fig:combined}. For instance, we improved the adversarial accuracy on Auto Attack by $39.63\%$ with a minimal drop of $1.58\%$ in the clean accuracy for ResNet-34 architecture on CIFAR-$10$ dataset. The combined performance for other attacks is provided in supplementary.

We compare our proposed method, DAD (no training data used) with existing state-of-the-art data-dependent approaches in Table~\ref{tab:soa}. We observe that our proposed method achieves decent adversarial accuracy in comparison to most of the data-dependent methods while maintaining a higher clean accuracy, entirely at test-time.

\begin{table}[htp]
\centering
\scalebox{0.81}{
\begin{tabular}{|c|c|c|c|}
\hline
\textbf{Method}                                                                        & \textbf{Data-Free} & \textbf{Clean} & \textbf{Auto Attack} \\ \hline \hline
Zhang \etal, 2019~\cite{zhang2019theoretically}                         & \xmark                  & 82.0           & 48.7                 \\ 
{\color[HTML]{000000} Sehwag \etal, 2021~\cite{sehwag2021improving}} &\xmark                  & 84.38          & 54.43                \\ 
{\color[HTML]{000000} Kundu \etal, 2020~\cite{kundu2021dnr}}  & \xmark                  & 87.32          & 40.41                \\ 
\cellcolor[HTML]{FEFEFE}{\color[HTML]{000000} Atzmon \etal, 2019~\cite{atzmon2019controlling}} &
 \xmark  &
  81.30 &
  40.22 \\ 
\cellcolor[HTML]{FEFEFE}{\color[HTML]{000000} \begin{tabular}[c]{@{}c@{}}Moosavi-Dezfooli \etal, \\ 2019~\cite{moosavi2019robustness}\end{tabular}} &
  \xmark  &
  83.11 &
  38.50 \\ \hline \hline
\begin{tabular}[c]{@{}c@{}}Baseline \\ (at test time w/o our framework)\end{tabular}   & -                  & 93.07          & 0                    \\ \hline
DAD (\textbf{Ours})                                                                             & \cmark               & 93.05          & 37.37                \\ \hline
\end{tabular}
}
\caption{Comparison of our \textbf{d}ata-free \textbf{a}dversarial \textbf{d}efense (DAD) with recent data-dependent approaches for resnet18 on CIFAR-$10$.}
\label{tab:soa}
\end{table}
\vspace{-0.1in}
\section{Conclusion}
\label{sec:conclusion}
We presented for the first time a complete test time detection and correction approach for adversarial robustness in absence of training data. We showed the performance of each of our proposed modules: detection and correction. The experimental results across adversarial attacks, datasets, and architectures show the efficacy of our method.  The combined module performance does not compromise much on the clean accuracy besides achieving significant improvement in adversarial accuracy, even against state-of-the-art Auto Attack. Our data-free method even gives quite competitive results in comparison to data-dependent approaches. Apart from these, there are other benefits associated with our proposed framework, as described below:
\begin{itemize}
\itemsep0em
\item Our detection module is independent of the correction module. Thus, any state-of-the-art classifier-based adversarial detector can be easily adopted on our source-free UDA-based adversarial detection framework. 
\item Any data-dependent detection approach can benefit from our correction module at test time to correct adversarial samples after successfully detecting them.
\end{itemize}
However, our adversarial detection method requires logits as input and hence is not strictly a black-box defense. Along with this, improving our radius selection algorithm to better estimate the optimal radius ($r^{*}$) are our future directions. 

\section{Acknowledgements}
\label{sec:Ack}
 This work is supported by a Start-up Research Grant (SRG) from SERB, DST, India (Project file number: SRG/2019/001938). 

{\small
\bibliographystyle{ieee_fullname}
\bibliography{references}
}
\newpage
\onecolumn
\begin{center}
    \LARGE{\textbf{{\textit{Supplementary for:} \\``\textbf{DAD}: \textbf{D}ata-free \textbf{A}dversarial \textbf{D}efense at Test Time"}}}

\end{center}

\setcounter{section}{0}
\setcounter{table}{0}
\setcounter{figure}{0}
\setcounter{equation}{0}
\vspace{18pt}
\hrule
\vspace{18pt}

\section{Qualitative Analysis of Low Frequency Component of adversarial data at different Radius}
\label{sec:visualization}
\vspace{0.1in}
\begin{figure*}[htp]
\centering
\centerline{\includegraphics[width=1.1\textwidth]{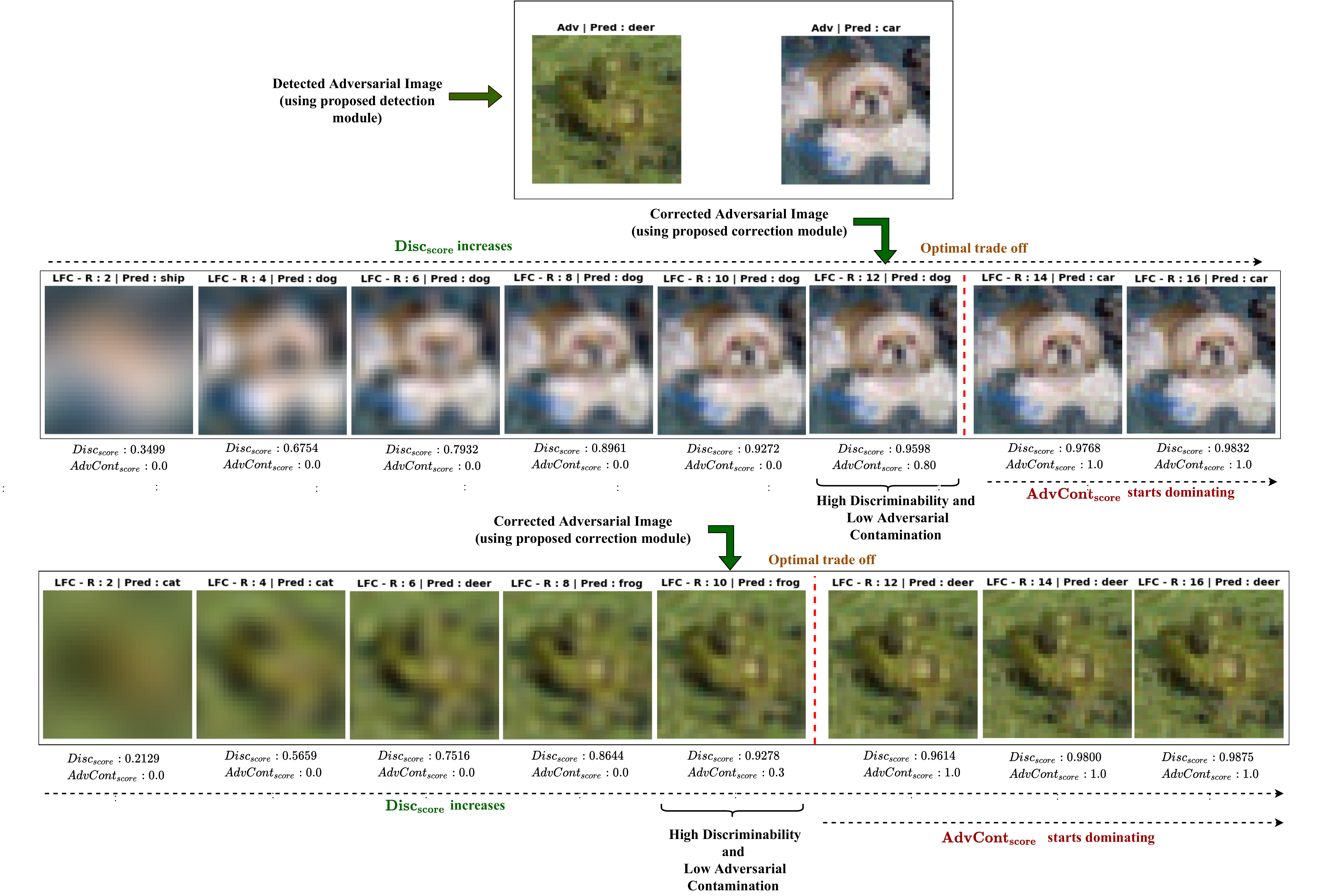}}
\vspace{0.3 in}
\caption{Visually demonstrating the trade off between discriminability and adversarial contamination. Our correction module suitably handles it through the proposed Algorithm $1$ in the main draft by selecting LFC in the spatial domain at optimal radius ($r^{*}$) having high $Disc_{score}$ and low $AdvCont_{score}$ i.e. max radius at which $Disc_{score}$ dominates over $AdvCont_{score}$. At radius ($r^{*}+1$), the adversarial contamination starts strongly influencing the predictions of pretrained model ($T_{m}$) and consequently $T_{m}$'s prediction for all subsequent radius remains same as the detected adversarial image's prediction. As shown, $r^{*}+1$ is $14$ and $12$ in the top and bottom row respectively. Thus, we select the radius($r^{*}$) as $12$ and $10$ in these cases and the corresponding LFC in the spatial domain when passed to the model ($T_{m}$) yields correct predictions. 
}
\label{fig:detailed_diagram}
\end{figure*}

\section{Performance of Proposed Detection Module using different Arbitrary Datasets}
As described in Sec. $4.1$ in the main draft, the target detector model $F_{t}$ is trained by adapting the source-detector model $F_{s}$. The model $F_{s}$ comprises of $S_{m}$ and $L_{advdet}$ where $S_{m}$ is trained on an arbitrary dataset $D_{arbitrary}$. Hence, to assess the effect of the choice of $D_{arbitrary}$ on the detection module and consequently our proposed method DAD (combined detection and correction), we conduct experiments on two distinct source datasets for each target dataset (i.e. CIFAR-$10$ and FMNIST) in addition to the results provided on TinyImageNet~\cite{le2015tiny} (as source dataset) in the Table $1$ in the main draft. We perform experiments with FMNIST and MNIST as $D_{arbitrary}$ for CIFAR-$10$, while MNIST and CIFAR-$10$ as $D_{arbitrary}$ for FMNIST. Similar to the results presented in Table $1$ in the main draft, we observe from Table~\ref{tab:arbitrary_detect} (shown below) that we achieve impressive detection accuracy across both the target datasets for each corresponding source dataset ($D_{arbitrary}$). 
\begin{table}[htp]
\centering
\scalebox{0.99}{
\begin{tabular}{|c|c|c|c|c|}
\hline 
{\color[HTML]{000000}Target Dataset }   & {\color[HTML]{000000} Source Dataset (Arbitrary Data)} & {\color[HTML]{000000} Detection Accuracy} & {\color[HTML]{000000} Clean Accuracy} & {\color[HTML]{000000} Adversarial Accuracy} \\ \hline \hline
\multirow{2}{*}{CIFAR-10} & MNIST    & 93.03 & 99.98 & 86.08 \\ 
                          & FMNIST  & 93.54 & 99.85 & 87.23 \\ \hline
\multirow{2}{*}{FMNIST}   & MNIST    & 88.51 & 99.03 & 77.99 \\ 
                          & CIFAR-10 & 84.41 & 84.07 & 84.75 \\ \hline
\end{tabular}
}
\caption{Results (in \%) of our proposed detection module comprising clean and adversarial detection accuracy along with overall detection accuracy on Auto Attack, are reported for different target datasets ($D_{test}$, i.e. CIFAR-10 and FMNIST). For each target dataset, we also vary the source dataset ($D_{arbitrary}$) which is completely different and arbitrary to the target dataset in terms of dissimilar semnatics and non-overlapping categories.}
\label{tab:arbitrary_detect}
\end{table}

\section{Combined (Detection and Correction) Performance on other attacks}
In order to evaluate the efficacy of our proposed approach (DAD) across a wide variety of attacks, we extend the analysis on combined performance presented in Sec. $6$ and Fig. $3$ of the main draft on the state-of-the-art Auto Attack to other popular attacks such as PGD and IFGSM. We observed from Table~\ref{tab:combined_other} that we achieve a respectable adversarial accuracy of more than $35\%$ on CIFAR-$10$ and more than $21\%$ on FMNIST across architectures (Resnet-$18$ and Resnet-$34$) on both the attacks while maintaining reasonable clean accuracy.

\begin{table}[htp]
\centering
\scalebox{0.99}{
\begin{tabular}{|c|c|c|c|c|c|}
\hline
\multicolumn{1}{|l|}{{\color[HTML]{000000} }} &
  \multicolumn{1}{l|}{{\color[HTML]{000000} }} &
  \multicolumn{2}{c|}{{\color[HTML]{000000} PGD}} &
  \multicolumn{2}{c|}{{\color[HTML]{000000} IFGSM}} \\ \cline{3-6} 
\multicolumn{1}{|l|}{\multirow{-2}{*}{{\color[HTML]{000000} Dataset}}} &
  \multicolumn{1}{l|}{\multirow{-2}{*}{{\color[HTML]{000000} Model}}} &
  {\color[HTML]{000000} Clean Accuracy} &
  {\color[HTML]{000000} Adversarial Accuracy} &
  {\color[HTML]{000000} Clean Accuracy} &
 {\color[HTML]{000000} Adversarial Accuracy} \\ \hline \hline
{\color[HTML]{000000} } &
  {\color[HTML]{000000} resnet18} &
  {\color[HTML]{000000} 89.01} &
  {\color[HTML]{000000} 36.38} &
  {\color[HTML]{000000} 85.49} &
  {\color[HTML]{000000} 35.44} \\ 
\multirow{-2}{*}{{\color[HTML]{000000} CIFAR-10}} &
  {\color[HTML]{000000} resnet34} &
  {\color[HTML]{000000} 88.28} &
  {\color[HTML]{000000} 37.92} &
  {\color[HTML]{000000} 88.61} &
  {\color[HTML]{000000} 31.94} \\ \hline
{\color[HTML]{000000} } &
  {\color[HTML]{000000} resnet18} &
  {\color[HTML]{000000} 90.09} &
  {\color[HTML]{000000} 21.29} &
  {\color[HTML]{000000} 88.17} &
  {\color[HTML]{000000} 23.15} \\ 
\multirow{-2}{*}{{\color[HTML]{000000} FMNIST}} &
  {\color[HTML]{000000} resnet34} &
  {\color[HTML]{000000} 90.45} &
  {\color[HTML]{000000} 21.79} &
  {\color[HTML]{000000} 90.57} &
  {\color[HTML]{000000} 22.21} \\ \hline
\end{tabular}
}
\caption{Performance of our proposed method (DAD) on PGD and IFGSM adversarial attacks where we report the overall clean and adversarial accuracy (in \%) across different architectures i.e. Resnet-$18$ and Resnet-$34$ for CIFAR-$10$ and FMNIST.}
\label{tab:combined_other}
\end{table}

\section{Combined (Detection and Correction) Performance on MNIST}
In this section, we evaluate our proposed combined module (detection module followed by correction module) solution strategy, i.e., DAD on the MNIST dataset~\cite{lecun1998gradient}, apart from FMNIST and CIFAR presented in Sec. $6$ (Figure $3$) of the main draft, to further validate our framework’s performance. 

We provide combined results on three distinct choices of $D_{arbitrary}$ i.e. CIFAR-$10$, FMNIST, and TinyImageNet for our target dataset MNIST ($D_{test}$). We observed (in Table~\ref{tab:combined_mnist}) that we achieve a significant boost in the adversarial accuracy across all three choices, without compromising much on the clean accuracy. These results verify that DAD can achieve good performance on a wide range of target datasets ($D_{test}$) for different choices of the source datasets ($D_{arbitrary}$). Please note the clean accuracy and adversarial accuracy of $T_{m}$ (resnet18) without our framework was $99.29\%$ and $0.00\%$ respectively against state-of-the-art Auto Attack.

\begin{table}[htp]
\centering
\scalebox{0.99}{
\begin{tabular}{|c|c|c|c|}
\hline
{\color[HTML]{000000} }             & {\color[HTML]{000000} }         & \multicolumn{2}{c|}{{\color[HTML]{000000} Auto Attack}}     \\ \cline{3-4} 
\multirow{-2}{*}{{\color[HTML]{000000} \begin{tabular}[c]{@{}c@{}}Source (arbitrary)\\ Dataset\end{tabular}}} &
  \multirow{-2}{*}{{\color[HTML]{000000} Model}} &
  {\color[HTML]{000000} Clean Accuracy} &
  {\color[HTML]{000000} Adversarial Accuracy} \\ \hline \hline
{\color[HTML]{000000} CIFAR-10}     & {\color[HTML]{000000} resnet18} & {\color[HTML]{000000} 90.46} & {\color[HTML]{000000} 34.77} \\ \hline
{\color[HTML]{000000} FMNIST}       & {\color[HTML]{000000} resnet18} & {\color[HTML]{000000} 96.12} & {\color[HTML]{000000} 31.15} \\ \hline
{\color[HTML]{000000} TinyImageNet} & {\color[HTML]{000000} resnet18} & {\color[HTML]{000000} 90.29} & {\color[HTML]{000000} 33.19} \\ \hline
\end{tabular}
}
\caption{Results (in \%) on MNIST using our proposed DAD framework containing detection and correction modules. In the detection module, the target detection model ($F_{t}$) is obtained by adapting the source detection model ($F_{s}$) using source-free UDA. The model $S_{m}$ is appended with detection layers to form $F_{s}$. So, we also report the performances for different choices of dataset $D_{arbitrary}$ (i.e. CIFAR-10, FMNIST and TinyImagenet) on which $S_{m}$ is trained.}
\label{tab:combined_mnist}
\end{table}
\pagebreak
\section{Attack Parameters for various arbitrary datasets}

\begin{table}[htp]
\centering
\scalebox{0.99}{
\begin{tabular}{|c|c|c|c|}
\hline
{\color[HTML]{000000} {Source/arbitrary datasets}} & $\epsilon$ & $\epsilon_{step}$ & Number of iterations \\ \hline \hline
TinyImageNet & 8/255 & 2/255 & 20  \\ \hline
MNIST        & 0.3   & 0.01  & 100 \\ \hline
CIFAR-10     & 8/255 & 2/255 & 20  \\ \hline
FMNIST       & 0.2   & 0.02  & 100 \\ \hline
\end{tabular}
}
\caption{The parameter values for PGD attack taken across different datasets for creating the adversarial dataset ($A_{arbitrary}$).}
\label{tab:my-table}
\end{table}

\section{Distribution of selected radius across samples}
As motivated in the Sec. $4.2$ of the main draft, our correction Algorithm $1$ estimates a suitable radius ($r^{*}$) entirely at the test time for each incoming sample without assuming any prior knowledge either about the training dataset or the adversarial attack. We demonstrated the importance of selecting $r^{*}$ optimally in Table $3$ of the main draft by comparing our correction algorithm’s performance with a random baseline (R.B.) (wherein a random radius is selected for each sample). We observed that although R.B.’s performance varied across datasets (being higher for CIFAR-$10$ than FMNIST), it was comfortably outperformed by our correction algorithm. 

\begin{figure}[htp]
\centering
\centerline{\includegraphics[width=1.0\textwidth]{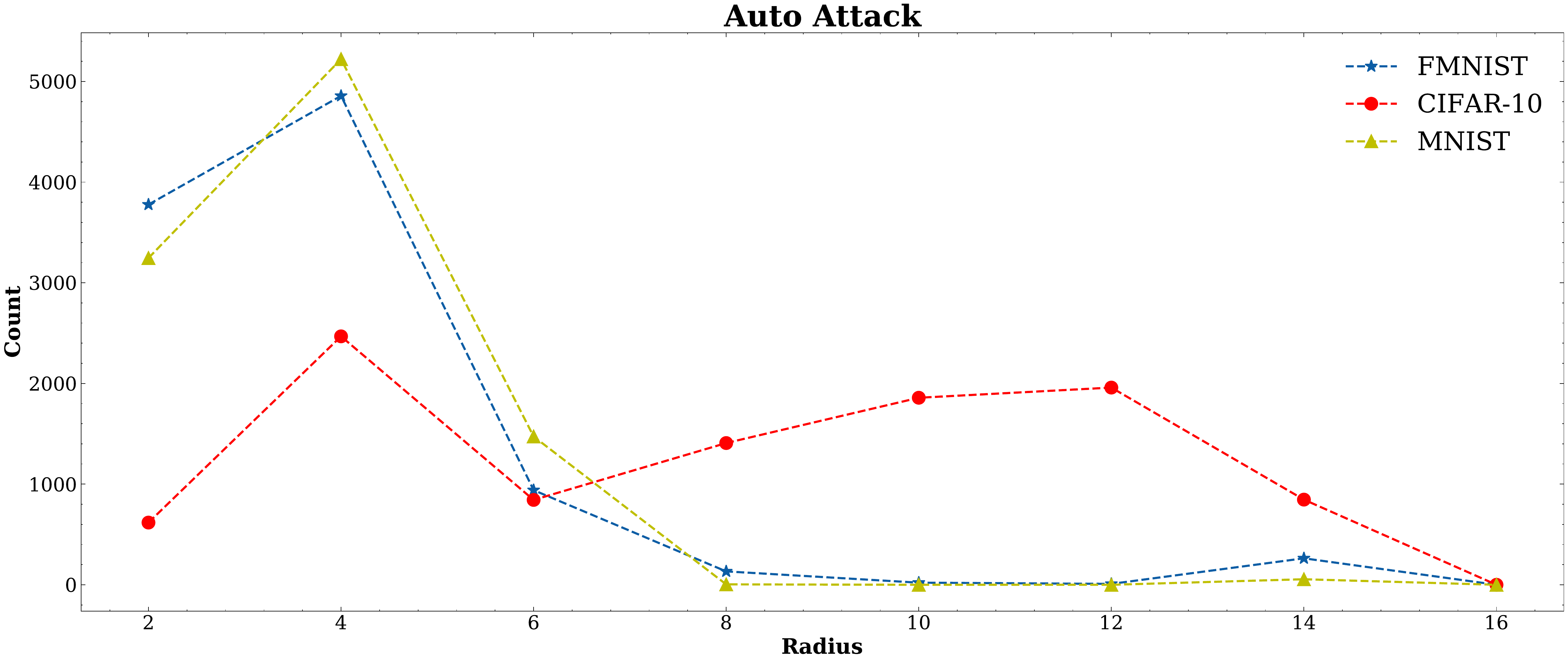}}
\caption{Distribution of radius selected by our proposed correction module on \textbf{Auto Attack} adversarial samples across different target datasets i.e. FMNIST, CIFAR-$10$ and MNIST.
}
\label{fig:radius_autoattack}
\end{figure}
\pagebreak
\begin{figure}[htp]
\centering
\centerline{\includegraphics[width=1.0\textwidth]{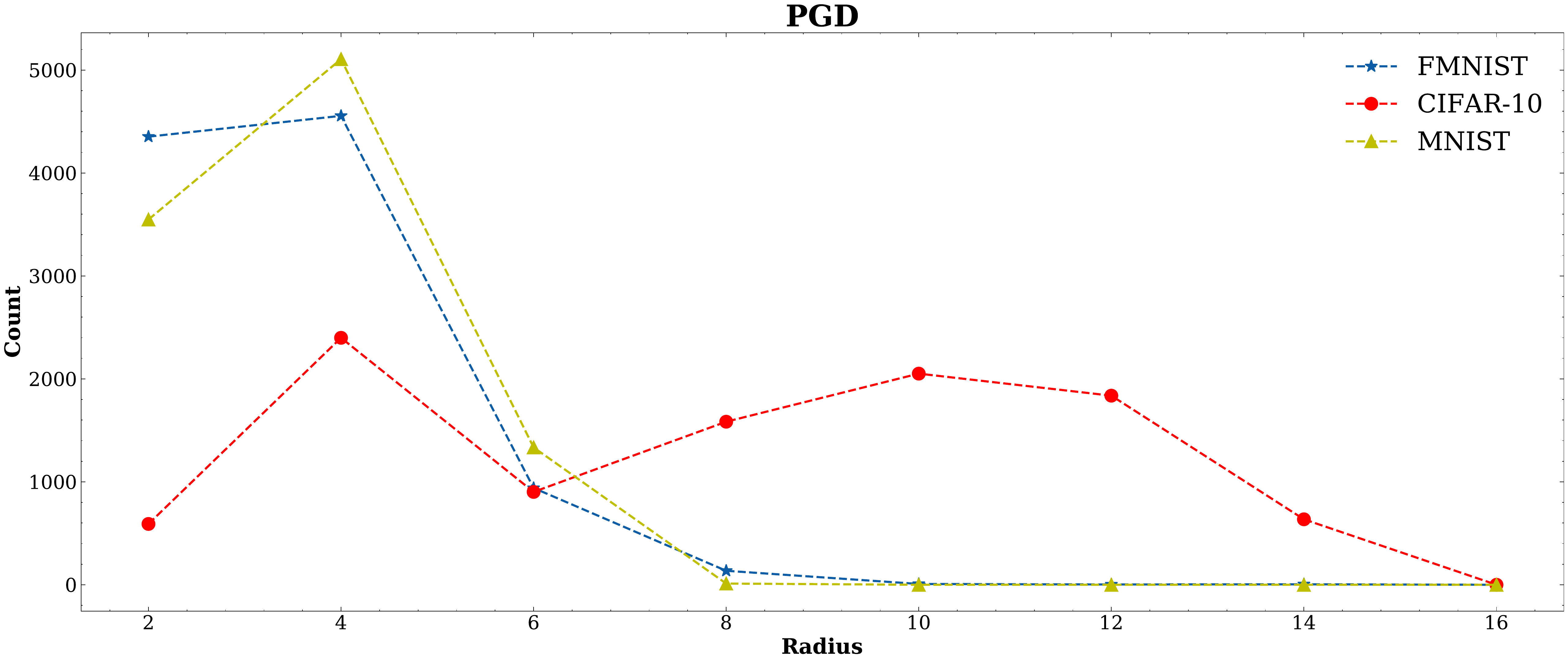}}
\caption{Distribution of radius selected by our proposed correction module on \textbf{PGD} Attack adversarial samples across different target datasets i.e. FMNIST, CIFAR-$10$ and MNIST.
}
\label{fig:radius_pgd}
\end{figure}
\vspace{-0.1in}
\begin{figure}[htp]
\centering
\centerline{\includegraphics[width=1.0\textwidth]{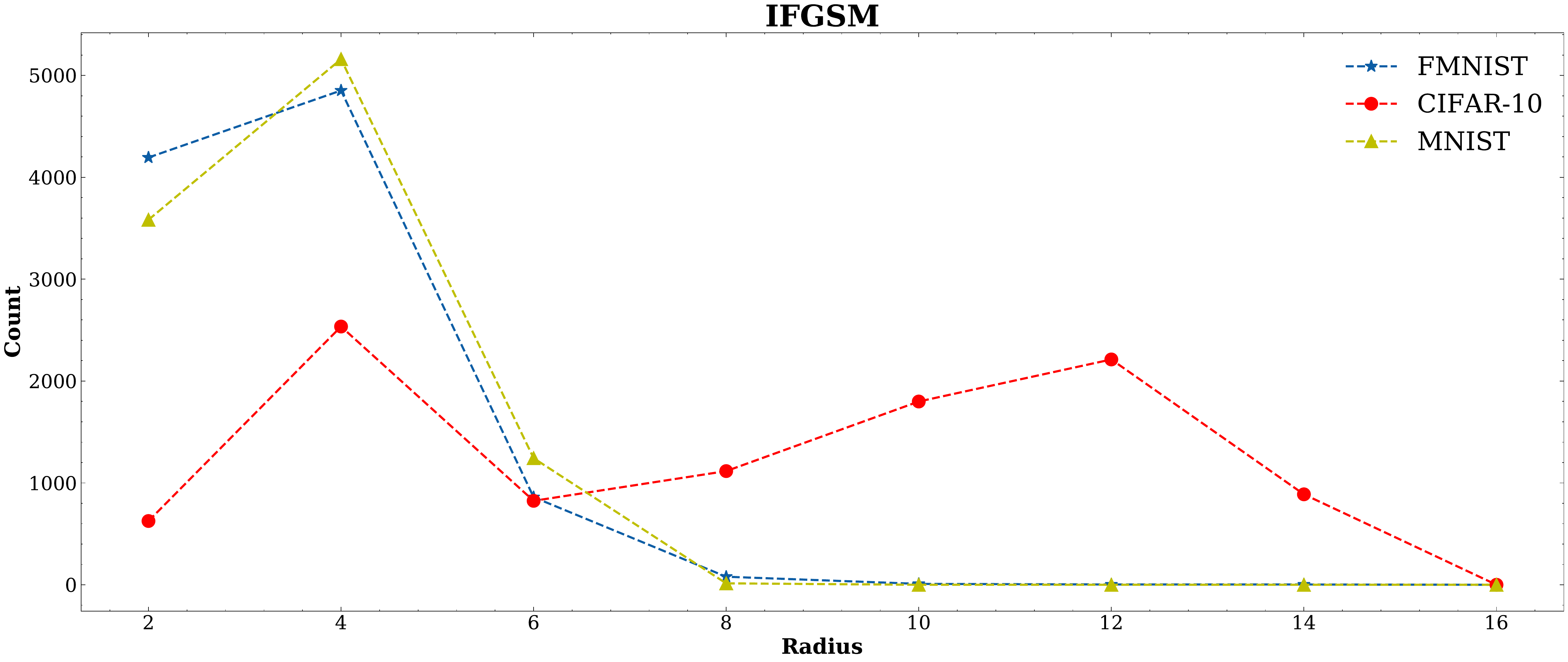}}
\caption{Distribution of radius selected by our proposed correction module on \textbf{IFGSM} Attack adversarial samples across different target datasets i.e. FMNIST, CIFAR-$10$ and MNIST.
}
\label{fig:radius_ifgsm}
\end{figure}

In order to investigate the selection of $r^{*}$ from another perspective, we plot the frequency distribution of $r^{*}$ for various attacks on Resnet-$18$ ($T_{m}$) trained on multiple datasets as shown in Figures~\ref{fig:radius_autoattack},~\ref{fig:radius_pgd} and ~\ref{fig:radius_ifgsm}. We observe that for MNIST and FMNIST datasets $r^{*}$ is majorly selected at a lower radius, whereas for the CIFAR-10 dataset the frequency distribution loosely resembles the uniform distribution. This explains the decent performance of R.B. on CIFAR-10 in Table $3$ of the main draft. More importantly, the figures indicate that $r^{*}$ can significantly vary across datasets (MNIST/FMNIST vs CIFAR-$10$). Moreover, $r^{*}$ can even vary across samples of a particular dataset (For \eg in CIFAR-10 samples, different radius are almost equi-proportionally selected). Our algorithm is able to accurately estimate $r^{*}$ on a sample-by-sample basis without any prior knowledge about the training dataset or the corresponding adversarial attack, as evident by the impressive performance shown in Table $2$ of the main draft.

We further compare our performance of a) carefully choosing a constant radius (for all the samples) with b) sample-level radius selection (finer-granularity control) i.e. $r^{*}$, through our proposed algorithm. We select the constant radius as $r=4$ since it’s the most frequently selected radius by our algorithm across different types of attacks and datasets (as shown in Figures~\ref{fig:radius_autoattack},~\ref{fig:radius_pgd} and ~\ref{fig:radius_ifgsm}). We observe that the sample-level selection strategy often provides significant improvements (shown in Table~\ref{tab:constant-radius}). For e.g. in CIFAR-$10$ we observe $\approx$ $16-18\%$ improvement by our algorithm over choosing a constant radius ($r=4$). However, we do notice on the MNIST dataset we observe slightly lower performance. Please note that our proposed algorithm obtains non-trivial improvement in adversarial accuracy even when choosing a constant radius or selecting radius at the sample level. We prefer the sample-level radius selection approach, as we often obtain a large gain in performance across a number of architecture, attack, and datasets configurations.

\begin{table}[htp]
\centering
\begin{tabular}{|c|c|c|c|}
\hline
Dataset &
  Attack &
  \begin{tabular}[c]{@{}c@{}}Performance (in \%)\\ ($r = 4$)\end{tabular} &
  \begin{tabular}[c]{@{}c@{}}Performance (in \%)\\ ($r^{*}$ selected at sample level)\end{tabular} \\ \hline \hline
\multirow{3}{*}{CIFAR-10} & pgd         & 22.08 & 39.39 \\ \cline{2-4} 
                          & ifgsm       & 22.26 & 38.49 \\ \cline{2-4} 
                          & auto attack & 22.37 & 40.25 \\ \hline \hline
\multirow{3}{*}{FMNIST}   & pgd         & 27.16 & 32.22 \\ \cline{2-4} 
                          & ifgsm       & 27.84 & 32.38  \\ \cline{2-4} 
                          & auto attack & 30.74 & 35.80 \\ \hline \hline
\multirow{3}{*}{MNIST}    & pgd         & 47.59 & 44.6  \\ \cline{2-4} 
                          & ifgsm       & 48.63 & 44.76 \\ \cline{2-4} 
                          & auto attack & 48.76 & 45.81 \\ \hline
\end{tabular}
\caption{Performance comparison of a) choosing a constant radius $r=4$ v/s b) sample-level selection strategy i.e. $r^{*}$}
\label{tab:constant-radius}
\end{table}

\vspace{18pt}
\hrule
\vspace{18pt}

\end{document}